\documentclass[conference]{IEEEtran}
\IEEEoverridecommandlockouts
\usepackage{cite}
\usepackage{amsmath,amssymb,amsfonts}
\usepackage{algorithmic}
\usepackage{graphicx}
\usepackage{textcomp}
\usepackage{xcolor}
\usepackage{booktabs}
\def\BibTeX{{\rm B\kern-.05em{\sc i\kern-.025em b}\kern-.08em
    T\kern-.1667em\lower.7ex\hbox{E}\kern-.125emX}}
\begin{document}

\title{Ensemble-Based Deepfake Detection using State-of-the-Art Models with Robust Cross-Dataset Generalisation 
}
\author{
\IEEEauthorblockN{
Haroon Wahab\IEEEauthorrefmark{1}, 
Hassan Ugail\IEEEauthorrefmark{1}, 
Ismail Lujain Jaleel\IEEEauthorrefmark{1}
}
\IEEEauthorblockA{\IEEEauthorrefmark{1}Centre for Visual Computing and Intelligent Systems, University of Bradford, Bradford, UK\\
m.h.wahab@bradford.ac.uk, h.ugail@bradford.ac.uk, lujainjaleel@icloud.com}
}

\maketitle

\begin{abstract}
Machine learning-based Deepfake detection models have achieved impressive results on benchmark datasets, yet their performance often deteriorates significantly when evaluated on out-of-distribution data. In this work, we investigate an ensemble-based approach for improving the generalization of deepfake detection systems across diverse datasets. Building on a recent open-source benchmark, we combine prediction probabilities from several state-of-the-art asymmetric models proposed at top venues. Our experiments span two distinct out-of-domain datasets and demonstrate that no single model consistently outperforms others across settings. In contrast, ensemble-based predictions provide more stable and reliable performance in all scenarios. Our results suggest that asymmetric ensembling offers a robust and scalable solution for real-world deepfake detection where prior knowledge of forgery type or quality is often unavailable.
\end{abstract}

\begin{IEEEkeywords}
Deepfake detection, ensemble models, cross-dataset generalization
\end{IEEEkeywords}

\section{Introduction}

Advances in generative artificial intelligence have enabled the synthesis of hyperrealistic media, most notably in the form of deepfakes. These techniques, particularly those involving facial manipulations—altering identity, expressions, or movements—pose a growing concern due to their misuse in misinformation campaigns, reputation attacks, political manipulation, and broader threats to digital trust and privacy~\cite{MUSTAK2023113368}.

While generative AI encompasses a wide array of domains, such as artwork, object synthesis, and scene generation, the primary focus of deepfake research remains centred on human faces. This emphasis stems from the fundamental role that facial identity plays in communication, verification, and media credibility. With recent progress in generative adversarial networks (GANs) and diffusion-based models, the synthetic content being produced is increasingly difficult to distinguish from genuine imagery~\cite{ABBAS2024124260}.

The improved quality of deepfakes, however, has made their detection substantially more difficult. As generative models continue to evolve, the distinction between real and fabricated media becomes increasingly subtle, perpetuating what is often described as an adversarial dynamic between generation and detection~\cite{laurier2024cat}. A core challenge in this domain is the poor generalization of detectors: while many techniques demonstrate strong performance on the datasets they are trained on, their effectiveness tends to deteriorate when applied to unseen distributions, especially those involving different manipulation styles or compression artifacts~\cite{pei2024deepfake, deepfakebench}. This presents a major barrier to the practical deployment of detection systems in the wild.

This study investigates the robustness and generalizability of several recent deepfake detectors when applied in a cross-dataset evaluation setting. We select six state-of-the-art (SOTA) models from a widely used benchmark suite and rely on their official implementations for experimentation. Our key research question is whether a single model can consistently maintain leading performance across datasets with differing characteristics or whether such performance is too context-dependent to allow for reliable selection.

To explore this, we evaluate an alternative strategy: aggregating predictions from multiple models through an ensemble approach. Rather than relying on any individual model, we combine their output confidences at the prediction level. Our results show that ensemble-based predictions tend to achieve competitive or superior performance across datasets, suggesting their utility as a more robust option for deployment in real-world, variable conditions.

\section{Deepfake Detection Models}
This study selects six deepfake detection models that encompass various design paradigms, including fundamental CNN-based detectors, spatial models that emphasize representation learning for forgery detection, and frequency-based methods that leverage features from the frequency domain for detection. The models considered are MesoInception-4~\cite{afchar2018mesonet}, Xception~\cite{ff++}, Core~\cite{ni2022core}, FFD~\cite{ffd}, SRM~\cite{srm}, and UCF~\cite{yan2023ucf}. These were selected based on their presence in recent benchmarks and the availability of reproducible, open-source implementations.

Each of these models is grounded in a unique architectural philosophy and detection rationale. While some prioritize texture-level artifacts or spatial inconsistencies, others are designed to uncover high-frequency traces or disentangled feature representations associated with manipulation. To assess how these differing approaches generalize across datasets with varying characteristics, we provide a short description of each model in the subsections that follow.

\subsection{MesoInception-4}
MesoInception-4 (naive detector) is one of the earliest dedicated deep learning models for facial forgery detection, proposed at a time when video compression posed significant challenges for detection accuracy~\cite{afchar2018mesonet}. It was designed with the aim of detecting face manipulations, such as Deepfakes and Face2Face, using a shallow yet effective architecture. The model builds on the original Meso-4 network by incorporating modified inception modules in the early layers. These modules combine 1×1 convolutions with dilated 3×3 convolutions to capture multi-scale spatial features while keeping the parameter count low. The architecture consists of four convolutional blocks followed by two dense layers for classification. Its mesoscopic design targets mid-level image features that are robust to compression artifacts, making it suitable for real-world compressed video settings.

\subsection{Xception}
Xception is a deep convolutional neural network originally developed for large-scale image classification but later adopted widely in deepfake detection tasks due to its strong performance in transfer learning~\cite{ff++}. The model is built entirely from depthwise separable convolutions, allowing for efficient computation while preserving representational power. Xception follows an Inception-style design without intermediate filter concatenation, instead using a streamlined architecture with repeated residual modules. It includes 36 convolutional layers structured into entry, middle, and exit flows. In the context of deepfake detection, Xception has proven effective at learning subtle visual features and artifacts introduced by manipulation, particularly when fine-tuned on forgery-specific datasets. Despite being more computationally intensive than lightweight alternatives, it remains a widely used benchmark due to its balanced performance across multiple datasets.

\subsection{CORE}
CORE (Consistent Representation Learning) is a deepfake detection framework that addresses the generalization problem by enforcing consistency in learned feature representations across various transformations~\cite{ni2022core}. The model builds on the intuition that genuine and forged faces should maintain distinguishable internal representations, even under augmentations such as Gaussian blur, JPEG compression, or spatial shifts. To this end, CORE employs a backbone network—typically ResNet-18—augmented with a projection head and a consistency loss that minimizes the distance between representations of the same image under different views. This consistency objective, combined with the standard classification loss, encourages the model to focus on manipulation-relevant features that persist across input variations. The result is improved robustness to distribution shifts, making CORE more effective in detecting forgeries across diverse manipulations.

\subsection{FFD}
FFD (Face Forgery Detection) is an attention-guided deepfake detection framework designed to enhance both the classification and localization of manipulated facial regions~\cite{ffd}. Built upon backbone networks like Xception or VGG, FFD introduces a novel attention-based layer that refines intermediate feature maps by focusing on regions likely to be manipulated. This attention map is integrated directly into the feature processing pipeline, effectively modulating the spatial focus of the network during training and inference.

The attention mechanism is trained under three supervision regimes—supervised, weakly supervised, and unsupervised—allowing flexibility in handling datasets with or without ground-truth manipulation masks. Two approaches for attention map generation are explored: (i) a Manipulation Appearance Model (MAM), which uses PCA-derived basis maps, and (ii) a simpler direct regression method. The final classification output benefits from these maps by masking irrelevant regions and emphasizing manipulated regions.

\subsection{SRM}
This study~\cite{srm} proposes a two-stream architecture that combines RGB features with high-frequency noise residuals extracted using Spatial Rich Model (SRM) filters. The model consists of three core modules:

\begin{itemize}
    \item \textbf{Multi-scale High-frequency Feature Extraction:} High-pass SRM filters are applied not only to the input images but also to low-level feature maps at multiple scales, enhancing the capture of subtle manipulation artifacts.

    \item \textbf{Residual Guided Spatial Attention (RSA):} Attention maps are generated from high-frequency residuals to guide the RGB stream toward spatial regions more likely to contain forgery traces. This helps mitigate texture overfitting by emphasizing manipulation-relevant areas.

    \item \textbf{Dual Cross-Modality Attention (DCMA):} This module models the interaction between the RGB and high-frequency streams using a self-attention mechanism. It computes cross-modal attention weights, allowing the network to leverage complementary signals from both modalities during feature refinement.
\end{itemize}

The extracted features from both streams are fused via attention-based operations, and the final classification is performed using AM-Softmax loss to improve class separation.

\subsection{UCF}
UCF (Uncovering Common Features)~\cite{yan2023ucf} is a generalizable deepfake detection framework designed to overcome two major generalization bottlenecks: overfitting to content (e.g., background, identity) and overfitting to method-specific forgery artefacts. Unlike many prior methods that address only one of these issues, UCF introduces a multi-task disentanglement framework that isolates common forgery features—those that are consistent across different generation methods—and uses only these for detection.

The model architecture is composed of an encoder, a decoder, and two classification heads. The encoder disentangles the input image into three feature components: content, specific forgery features, and common forgery features. These features are learned using two distinct encoders (content and fingerprint) without shared parameters. The decoder reconstructs images using a conditional design via Adaptive Instance Normalization (AdaIN), blending content and forgery features. This ensures disentangled representations and helps preserve semantic consistency in reconstruction.

Two classification heads are trained in parallel: one predicts the specific forgery method (multi-class) using the specific forgery features, and the other performs binary classification (real vs. fake) using only the common forgery features. A contrastive regularization loss further encourages separation between real and fake representations and between different forgery types, improving feature discrimination. Additionally, cross-reconstruction and self-reconstruction losses ensure that the disentangled representations can accurately regenerate the input, further reinforcing the decoupling.

\begin{table*}[ht]
\centering
\caption{AUROC and AUPRC scores of individual models and ensemble variants on two out-of-domain datasets: UADFV and Celeb-DF-v2.}
\label{tab:performance}
\begin{tabular}{lcccc}
\toprule
\textbf{Model} & \textbf{AUROC (UADFV)} & \textbf{AUPRC (UADFV)} & \textbf{AUROC (Celeb-DF-v2)} & \textbf{AUPRC (Celeb-DF-v2)} \\
\midrule
MesoInception   &0.823                      &0.810                      &0.650                         &0.757                         \\
Xception        &0.936                      &0.943                      &0.740                         &0.837                         \\
Core            &\textbf{0.961}       &\textbf{0.964}                      &0.741                         &0.829                         \\
FFD             &0.950                      &0.948                      &0.687                         &0.785                         \\
SRM             &0.880                      &0.828                      &0.760                         &0.846                         \\
UCF             &0.920                      &0.924                   &\textbf{0.772}                         &\textbf{0.861}                         \\
Ensemble-Avg    &0.958                      &0.959                      &0.768                         &0.852                         \\
Ensemble-Weighted &0.958                    &0.960                      &0.769                         &0.853                         \\
\bottomrule
\end{tabular}
\end{table*}

\section{Model Ensembling for Robust Detection}

In this study, we adopt a \textit{probability-level ensembling strategy (late fusion)} to combine outputs from six diverse deepfake detection models. Each model produces a class probability score indicating the likelihood that a given input is manipulated. These models are \textit{asymmetric} in design, varying in architecture, feature extraction mechanisms, loss functions, and training inductive biases. This structural diversity is critical, as it has been empirically shown that \textit{asymmetric deep ensembles benefit more from increased ensemble size} than symmetric ones. Notably, a recent study demonstrated that ``deep ensembles built on asymmetric neural networks achieve significantly better performance as ensemble size increases compared to their symmetric counterparts''~\cite{chernov2025empirical}.

We evaluate two ensemble variants: (i) \textit{skill-weighted probability averaging}, and (ii) \textit{simple unweighted averaging}. Both operate at the soft probability level, combining the class probabilities predicted by each model rather than raw logits or final class labels.

Let:

\begin{itemize}
    \item $M = \{M_1, M_2, \dots, M_6\}$ denote the set of six models.
    \item $p_i(x) \in [0,1]$ be the predicted probability (of the input $x$ being fake) by model $M_i$.
    \item $w_i \in [0,1]$ be the assigned skill-based weight for model $M_i$, such that $\sum_{i=1}^{6} w_i = 1$.
\end{itemize}

\subsection{Skill-Weighted Probability Averaging}

In this strategy, models are assigned weights based on their individual performance (e.g., accuracy or AUC on a validation set). The final prediction $P_{\text{weighted}}(x)$ is computed as:

\begin{equation}
P_{\text{weighted}}(x) = \sum_{i=1}^{6} w_i \cdot p_i(x)
\end{equation}

Higher $w_i$ values indicate greater confidence or reliability of model $M_i$ based on validation performance. This allows more accurate models to have greater influence on the ensemble decision.

\subsection{Simple Probability Averaging}

In the absence of prior performance knowledge or under the assumption of equal reliability, we apply uniform weights. The final prediction $P_{\text{avg}}(x)$ is:

\begin{equation}
P_{\text{avg}}(x) = \frac{1}{6} \sum_{i=1}^{6} p_i(x)
\end{equation}

This method assumes all models contribute equally and reduces variance through averaging, which is theoretically justified under the bias–variance decomposition when models are independently trained.

These ensemble strategies offer a balance between simplicity and adaptability. The weighted ensemble captures model skill differences, while the unweighted version offers robustness and interpretability. We compare both in our experiments to assess their impact on cross-dataset generalization and detection accuracy.

\section{Experimental Setup}
FaceForensics (FF++)~\cite{ff++} is considered as a representative dataset covering diverse set of manipulations such as Deepfakes, Face2Face~\cite{thies2016face2face}, FaceSwap, NeuralTextures~\cite{thies2019deferred}. Therefore, all models are trained on the FF++ dataset. We use the model implementations and pretrained weights provided by the deepfake benchmark study~\cite{deepfakebench}. For evaluation, we test both the individual models and their ensemble variants on two out-of-domain datasets: Celeb-DF-v2 and UADFV.

For preprocessing, we apply a standardized pipeline comprising three steps: face detection, face cropping, and face alignment. All three steps are performed using the DLIB toolkit~\cite{dlib09}. The aligned face images are then resized to a fixed resolution of 256×256 pixels.

For evaluation, we report two standard classification metrics widely used in deepfake detection tasks: the Area Under the Receiver Operating Characteristic curve (AUROC) and the Area Under the Precision-Recall Curve (AUPRC).

\section{Results and Discussion}

\begin{figure*}[ht]
    \centering
    \includegraphics[width=0.48\linewidth]{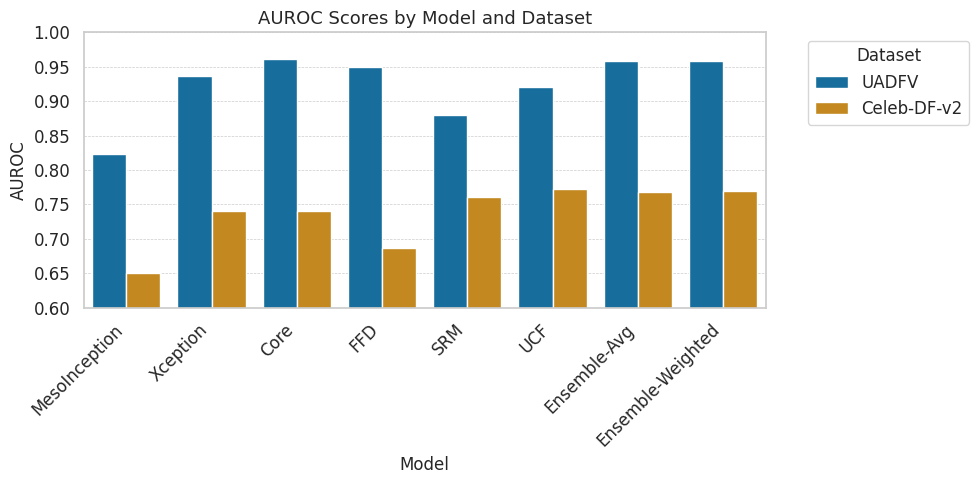}
    \includegraphics[width=0.48\linewidth]{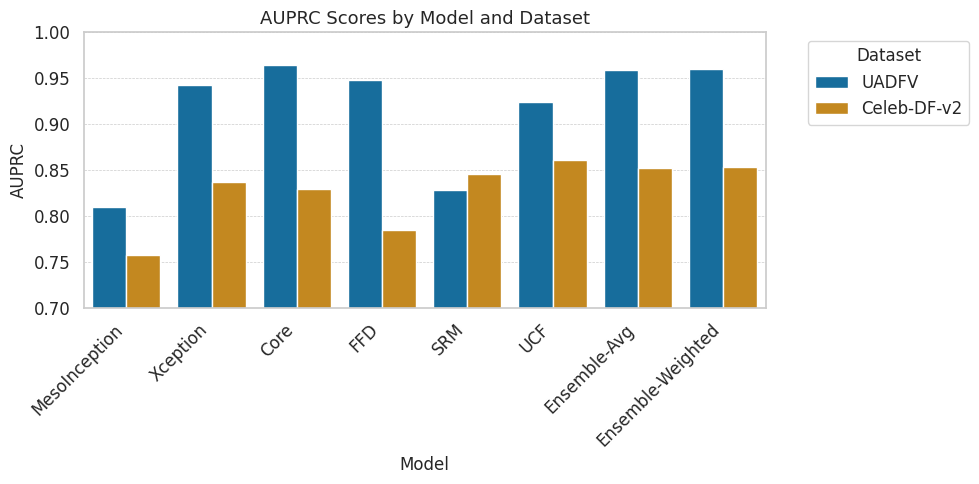}
    \caption{Comparison of AUROC (left) and AUPRC (right) for individual deepfake detectors and their ensemble combinations, evaluated on UADFV and Celeb-DF-v2 datasets.}
    \label{fig:barplots}
\end{figure*}

\begin{figure}[ht]
    \centering
    \includegraphics[width=0.48\textwidth]{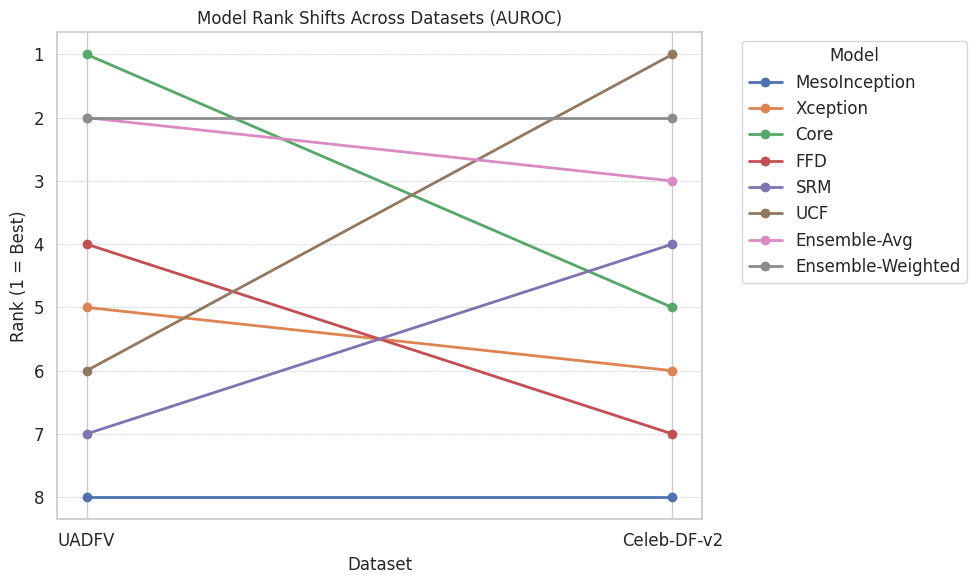}
    \caption{AUROC-based rank shifts across datasets. A lower position (1) indicates a better rank.}
    \label{fig:bumpchart}
\end{figure}

Table~\ref{tab:performance} presents the AUROC and AUPRC scores of all six individual detection models and the two ensemble strategies on the UADFV and Celeb-DF-v2 datasets. These results reveal clear performance differences across datasets and provide several insights into model generalization behavior and the stability of ensemble-based predictions.

A notable observation is the \textit{variance in model performance across datasets}. Several models that perform well on UADFV experience a significant drop when evaluated on Celeb-DF-v2. For example, the CORE achieves the highest AUROC score (0.961) on UADFV but performs much lower on Celeb-DF-v2 (AUROC = 0.741). Similarly, FFD drops from 0.950 AUROC on UADFV to 0.687 on Celeb-DF-v2. This indicates a lack of consistent generalization across data domains.

These results also highlight a \textit{shift in performance rankings} depending on the dataset. On UADFV, Core, FFD, and the ensemble variants perform near the top. In contrast, on Celeb-DF-v2, UCF achieves the highest AUROC (0.772), while Core and FFD fall behind. Such variability reinforces that no single model consistently dominates across datasets, complicating the choice of a “best” model in real-world, unseen scenarios.

In contrast, the ensemble methods—both averaging and skill-weighted—exhibit \textit{stable and near-optimal performance} on both datasets. On UADFV, their AUROC scores (0.958) are nearly equal to the best individual model (CORE), and on Celeb-DF-v2, they closely trail the top performer (UCF). Notably, ensembles never perform the worst in any setting, suggesting that they offer a reliable fallback when model selection is uncertain.

Figure~\ref{fig:barplots} provides a visual comparison of AUROC and AUPRC scores across both datasets. From the left panel, it is evident that CORE, FFD, and the ensemble methods perform strongly on UADFV, whereas their performance degrades substantially on Celeb-DF-v2. Meanwhile, UCF improves its standing on Celeb-DF-v2, highlighting the domain sensitivity of different models. The right plot (AUPRC) confirms the same general trend, that performance rankings are not preserved across datasets.

To further highlight the rank instability of individual models, Figure~\ref{fig:bumpchart} shows a bump chart based on AUROC scores. Models such as Core and FFD, which rank at the top on UADFV, experience a steep drop in rank on Celeb-DF-v2. Conversely, UCF rises from a mid-tier position to the top. The ensemble models, however, remain consistently near the top across both datasets with minimal rank fluctuation. This confirms their robustness and supports the claim that deep ensembles are a more reliable choice when facing unknown or shifting data distributions.

These results indicate that while individual models can be highly effective on specific datasets, their generalization ability varies significantly. In contrast, ensemble methods show stable performance and minimal variance in both score and rank, making them better suited for deployment in real-world scenarios where dataset characteristics are not known a priori.

\section{Conclusion}
In this work, we examined the robustness of recent state-of-the-art deepfake detection models under cross-dataset evaluation using two challenging out-of-domain benchmarks. Our analysis demonstrated that while individual models often perform well within the dataset they were trained on, their effectiveness tends to vary substantially when applied to unfamiliar data distributions. This inconsistency underscores the ongoing challenge of generalization in face forgery detection.

To address this issue, we explored ensemble-based prediction strategies, combining outputs from diverse detection models through probability-level fusion. Empirical results across AUROC and AUPRC metrics showed that both simple averaging and skill-weighted ensembles consistently maintained performance close to the best individual models. In particular, the ensembles never ranked the lowest in either dataset, reinforcing their potential as a robust fallback in uncertain real-world settings.

The stability of ensemble performance across datasets is promising: when no prior information is available about the nature of the manipulation or the domain characteristics, ensemble methods offer a dependable baseline. In real-world deployments, where unseen manipulations and data shifts are the norm, this resilience makes deep ensembles a compelling alternative to any single-model solution. Further exploration could include dynamic ensemble weighting and model selection conditioned on content characteristics.


\end{document}